\pdfoutput=1

\documentclass[11pt]{article}

\usepackage[final]{acl}

\usepackage{times}
\usepackage{latexsym}

\usepackage[T1]{fontenc}

\usepackage[utf8]{inputenc}

\usepackage{microtype}

\usepackage{inconsolata}

\usepackage{graphicx}

\usepackage{bm}
\usepackage{amsmath}
\usepackage{amsfonts}
\usepackage{booktabs}
\usepackage{multirow}
\usepackage{enumitem}
\usepackage{subcaption}
\usepackage{flushend}
\usepackage{algorithm2e}
\RestyleAlgo{ruled}
\usepackage{colortbl}

\usepackage{amsmath,amsfonts,bm}

\def\eqref#1{equation~\ref{#1}}

\def\1{\bm{1}}

\def\vr{{\bm{r}}}

\def\vz{{\bm{z}}}

\def\mE{{\bm{E}}}

\DeclareMathAlphabet{\mathsfit}{\encodingdefault}{\sfdefault}{m}{sl}
\SetMathAlphabet{\mathsfit}{bold}{\encodingdefault}{\sfdefault}{bx}{n}

\def\gE{{\mathcal{E}}}

\def\gG{{\mathcal{G}}}

\def\gP{{\mathcal{P}}}
\def\gQ{{\mathcal{Q}}}

\def\gS{{\mathcal{S}}}

\def\gV{{\mathcal{V}}}

\newcommand{\OurName}{\texttt{PECAN}}

\title{PECAN: LLM-Guided Dynamic Progress Control with Attention-Guided Hierarchical Weighted Graph for Long-Document QA}

\author{
Xinyu Wang\textsuperscript{\rm1,2}, 
Yanzheng Xiang\textsuperscript{\rm2}, 
Lin Gui\textsuperscript{\rm2}, 
Yulan He\textsuperscript{\rm1,2,3} \\
\textsuperscript{1}Department of Computer Science, University of Warwick \\
\textsuperscript{2}Department of Informatics, King's College London\\
\textsuperscript{3}The Alan Turing Institute\\
\texttt{Xinyu.Wang.11@warwick.ac.uk} \\
\texttt{\{yanzheng.xiang, lin.1.gui, yulan.he\}@kcl.ac.uk}
}

\begin{document}

\maketitle

\begin{abstract}

Long-document QA presents challenges with large-scale text and long-distance dependencies. 
Recent advances in Large Language Models (LLMs) enable entire documents to be processed in a single pass. However, their computational cost is significantly high. 
Retrieval-Augmented Generation (RAG) methods split text into smaller chunks, but they often yield inferior results and may lose global context. 
Recent approaches that integrate LLMs into RAG via iterative summarization either underutilize LLM capabilities or still incur high computational costs.
In this paper, we combine the high accuracy of LLMs with the efficiency of RAG and propose LLM-Guided Dynamic \underline{P}rogr\underline{e}ss \underline{C}ontrol with \underline{A}ttentio\underline{n}-Based Hierarchical Weighted Graph (\OurName). 
Our method introduces two key improvements:
(1) LLM-Guided Dynamic Progress Control: We leverage LLMs to dynamically control the retrieval process, adjusting the amount of retrieved information based on different queries to achieve a better balance of effectiveness and efficiency.
(2) Attention-Guided Retrieval: We propose a novel retrieval method that constructs a hierarchical graph where edges are derived by LLM attention weights.
Experimental results demonstrate that \OurName\ achieves LLM-level performance while maintaining computational complexity comparable to that of RAG methods on two single-document and two multi-document QA datasets.
\footnote{Code is available at \url{https://github.com/xnyuwg/pecan}.}

\end{abstract}


\section{Introduction}

\begin{figure*}[t]
	\centering 
	\centerline{\includegraphics[width=0.99\textwidth]{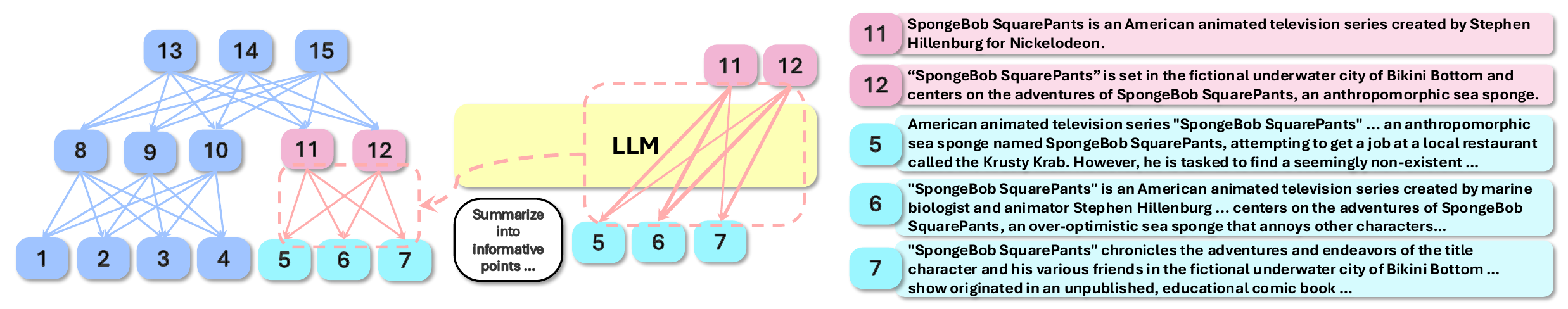}}
	\caption{
            Overview of the attention-guided many-to-many summarization process for constructing the Hierarchical Weighted Directed Acyclic Graph (HWDAG).
            Each node represents an \textbf{Information Point (IP)}, typically summarizing one or a few events, and can have multiple predecessors and successors. 
            The edge weights, derived from the LLM's attention during summarization, indicate the strength of relationships 
            and the flow of information from lower-level nodes to higher-level nodes. 
            The thickness of the red line on the right indicates the magnitude of the attention weights.
            For example, node \#11 is more closely related to nodes \#5 and \#6, while node \#12 is more associated with nodes \#6 and \#7. In this instance, the LLM extracts two events from three text chunks, with node \#6 referencing both events. 
            (For brevity, long text are truncated.)
        }
	\label{fig:sum} 
\end{figure*}

Long-document Question Answering (QA) poses several challenges, including handling large-scale texts,
uneven information distribution, and long-distance dependencies.
While Large Language Models (LLMs) \citep{
dubey2024llama3herdmodels,yang2024qwen2technicalreport,geminiteam2024geminifamilyhighlycapable,openai2024gpt4technicalreport} can process entire documents after undergoing a context extension training stage, this approach is computationally intensive and inefficient, as 
queries often target only small segments of the text. 
In contrast, Retrieval-Augmented Generation (RAG) 
\citep{
tay2022transformer,santhanam-etal-2022-colbertv2,lin-etal-2023-train}
mitigate this by segmenting long documents into chunks and retrieving the most relevant segments. 
However, studies like LongBench \citep{bai-etal-2024-longbench} have demonstrated that the performance of RAG methods is often inferior to directly feeding the full document into LLMs \citep{zhang-etal-2023-extractive,nair-etal-2023-neural,newman-etal-2023-question}.
RAG methods may suffer from incomplete retrieval, 
lack of global context, and struggle to capture long-distance dependencies.
Recent methods, such as MeMWalker \citep{chen2023walkingmemorymazecontext} and RAPTOR \citep{sarthi2024raptor} employ LLMs to iteratively summarize text into a tree.
These methods use concise summaries generated by LLM to capture the global context while leveraging RAG to retrieve detailed information.  
However, MeMWalker still incurs a high computational cost since it repeatedly relies on the LLM to determine retrieval strategies, while RAPTOR only uses the LLM for summarization and heavily depends on traditional RAG. 

In this paper,
we propose a new method, LLM-Guided Dynamic \underline{P}rogr\underline{e}ss \underline{C}ontrol with \underline{A}ttentio\underline{n}-Based Hierarchical Weighted Graph (\textbf{\OurName}), combining LLM accuracy and RAG efficiency. Unlike existing approaches, our method dynamically structures and searches information using an \textbf{Attention-Guided Hierarchical Graph}.
Our approach comprises two stages: \emph{Attention Graph Construction} and \emph{Dynamic Graph Search}. In the \emph{Attention Graph Construction} stage, 
we 
prompt an LLM to generate multiple \textbf{Information Points (IPs)}, each typically focusing on a single or a few events. These IPs form a Hierarchical Weighted Directed Acyclic Graph (HWDAG), where LLM attention weights between the generated nodes and the input nodes define relationships between lower- and higher-level nodes.
This structure efficiently consolidate and summarize information about the same event from various sources, and enabling efficient event identification, as illustrated in Figure~\ref{fig:sum}.

Building on this graph, during the \emph{Dynamic Graph Search} stage, instead of retrieving fixed-size text chunks, we use LLM attention to identify relevant high-level summaries and trace back to detailed information.
This method effectively utilizes the LLM’s knowledge, capturing both the generated summarization and the dynamic flow of information between high-level summaries and their corresponding detailed content within the graph.
In addition, previous RAG methods retrieve a static number of top-ranked text chunks, 
which fails to account for the fact that different queries may require varying amounts of information. 
This fixed setting might lead to computational waste for simple queries and insufficient retrieval for more complex ones. 
We introduce a novel method called \textbf{Dynamic Progress Control}, 
which allows the LLM to adaptively determine the amount of information needed per query. 
This approach essentially reallocates computational resources based on the LLM’s knowledge to achieve a better balance of effectiveness and efficiency.
Meanwhile, previous inputs are KV-cached so that new documents can be appended without reprocessing the entire input. 
As a result, our method can effectively handle queries that require varying amounts of information, particularly when the necessary details are distributed across different parts of the graph. 
This avoids additional computational overhead and resolves the challenge of determining the optimal number of chunks to retrieve.

In experiments, \OurName\ achieves a good balance between effectiveness and efficiency, attaining LLM-level performance while keeping computational costs on par with RAG methods.

In summary, our contributions in this paper include:
\begin{itemize}[noitemsep,nolistsep,leftmargin=*]
\item 
We leverage LLMs to dynamically control the retrieval process, adjusting the amount of retrieved information per query without incurring additional computational overhead, leading to a more balanced effectiveness and efficiency.
\item 
We propose a novel attention-guided retrieval paradigm that constructs a many-to-many graph using LLM attention weights. In this graph, edges are derived from attention weights, and retrieval is guided accordingly. Each node represents an Information Point (IP) focusing on one or a few events, allowing for structured event tracking rather than simple text chunk retrieval. 
\item 
Our empirical results show that \OurName\ achieves LLM-level accuracy while maintaining computational efficiency of traditional RAG methods.
\end{itemize}


\section{Related Work}

\paragraph{Long-Context Language Models}
Recent long-context language models have focused on overcoming fixed context window limitations, primarily through positional interpolation and training on full-length texts.
\citet{chen2023extendingcontextwindowlarge,peng2024yarn,fu2024dataengineeringscalinglanguage} fine-tuned models on longer inputs and extended Rotary Position Embedding (RoPE; \citealp{su2023roformerenhancedtransformerrotary}) for extended contexts.
LongRoPE \citep{ding2024longrope} performs direct extrapolation by rescaling RoPE 
with varied interpolation.
LongAlign \citep{bai2024longalign} constructs a long-context dataset, adopting packing and sorted batching strategies.
PoSE \citep{zhu2024pose} manipulates position indices by skipping bias terms.
SkipAlign \citep{wu2024longcontextalignmentshort} synthesizes long-range dependencies from the aspect of position indices.
Infini-Transformer \citep{munkhdalai2024leavecontextbehindefficient} handles infinitely long inputs using compressive memory, masked local attention, and long-term attention mechanisms.
Our primary focus is on effectively leveraging LLM capabilities, and the LLMs employed in these techniques can serve as the base models in our framework.

\paragraph{RAG}
Traditional retrieval techniques, such as TF-IDF \citep{tfidf1972} and BM25 \citep{robertson1995okapi,bm252009}, rely on word-term matching.
Subsequently, deep learning–based retrieval methods quickly gained popularity \citep{realm2020, min-etal-2021-joint, izacard2022atlasfewshotlearningretrieval, izacard2021distilling, wang-etal-2023-simlm}.
Among these, 
DPR \citep{karpukhin-etal-2020-dense} encodes queries and documents as dense embeddings.
ColBERT \citep{ColBERT2020,santhanam-etal-2022-colbertv2} produces multi-vector representations.
DHR \citep{liu-etal-2021-dense-hierarchical} leverages both document-level and passage-level semantics.
CPT-text \citep{neelakantan2022textcodeembeddingscontrastive} utilizes contrastive pre-training on unsupervised data.
NCI \citep{wang2022a} directly generates relevant document identifiers.
RETRO \citep{borgeaud2022improvinglanguagemodelsretrieving,wang-etal-2023-shall} conditions on document chunks based on local similarity.
HHR \citep{arivazhagan-etal-2023-hybrid} combines sparse and dense retrieval methods.
Dragon \citep{lin-etal-2023-train} uses contrastive learning and data augmentation to train a model, achieving state-of-the-art retrieval performance. 

\paragraph{LLM and RAG}
Additionally, with the rise of LLMs, several studies have explored combining LLMs with RAG.
GENREAD \citep{yu2023generate} prompts LLMs to generate contextual documents.
RECITE \citep{sun2023recitationaugmented} retrieves from the LLM’s internal memory.
KGP \citep{wang2023knowledgegraphpromptingmultidocument} builds a knowledge graph with the LLM navigating.
Recently,
MeMWalker \citep{chen2023walkingmemorymazecontext} constructs tree-based summaries and uses LLMs to navigate.
RAPTOR \citep{sarthi2024raptor} creates tree-based summaries and leverages embedding similarities to select the most relevant nodes at each level for retrieval.
In contrast, our approach makes greater use of LLM knowledge by employing attention mechanisms to construct a graph and perform the search, allowing navigation along multiple paths and termination at any depth.


\section{Methodology}
\label{sec_method}

\begin{figure*}[t]
	\centering 
	\centerline{\includegraphics[width=0.99\linewidth]{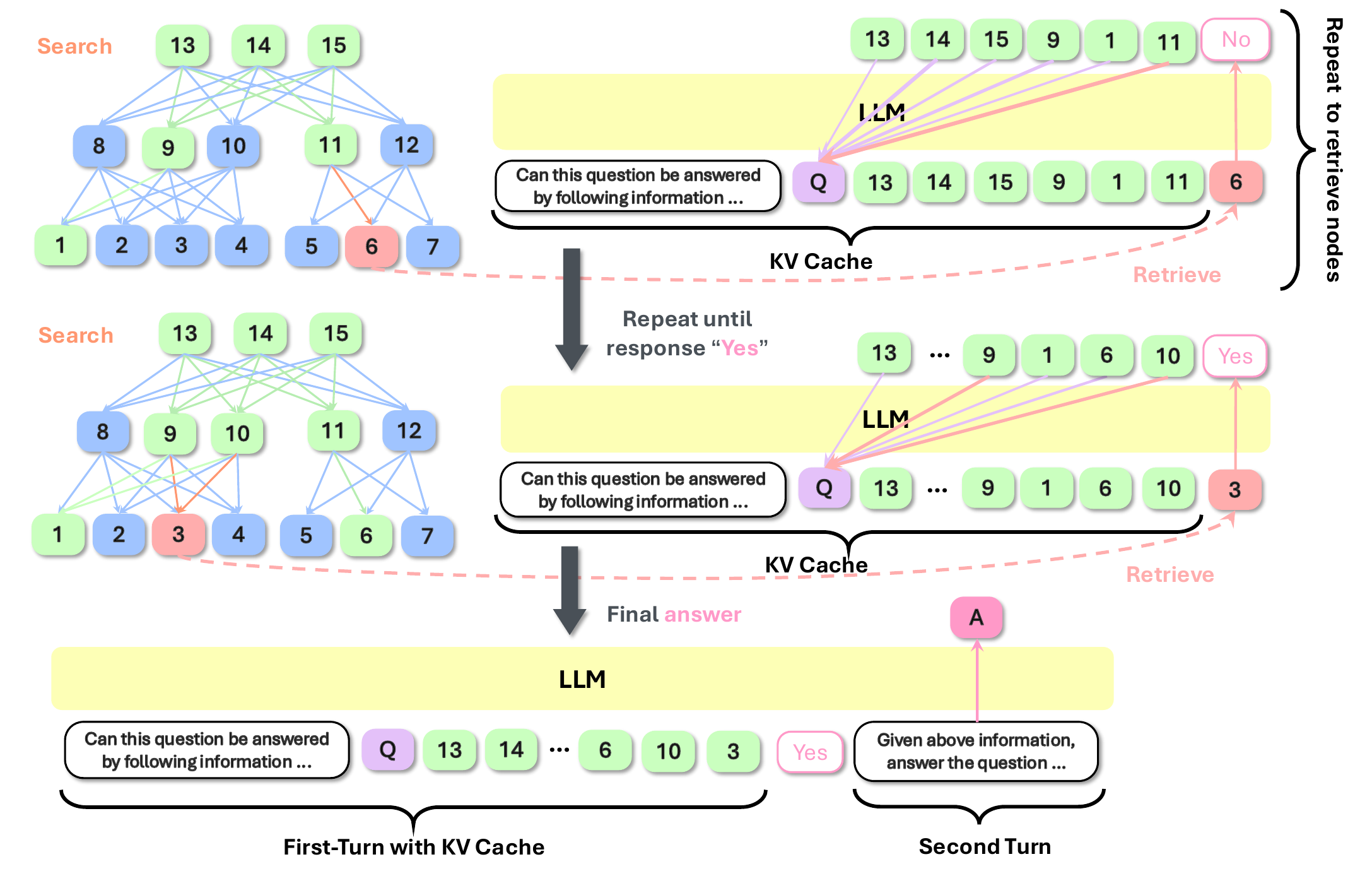}}
	\caption{
            Overview of \emph{Dynamic Graph Search}. 
            At each iteration, a node is retrieved based on attention weights when searching the graph.
            The visited nodes are then fed into the LLM, prompting it to determine whether sufficient nodes have been gathered to answer the query.
            Due to KV caching, this process adds no additional computational cost.
            The search continues until the LLM indicates that enough relevant nodes have been retrieved, at which point the final answer is generated.
            This procedure dynamically adapts to the query, retrieving nodes flexibly across multiple graph paths and depths.
        }
	\label{fig:sea} 
\end{figure*}

Our method consists of two main steps: \emph{Attention Graph Construction} and \emph{Dynamic Graph Search}. 
In the \emph{Attention Graph Construction} stage, illustrated in Figure~\ref{fig:sum}, we utilize the LLM's attention weights to build a Hierarchical Weighted Directed Acyclic Graph (HWDAG) from documents.
This is a one-time preprocessing step for each document, after which it can be reused for any query to that document.
In the \emph{Dynamic Graph Search} stage, as illustrated in Figure~\ref{fig:sea}, we dynamically control the volume of retrieved nodes and perform a search guided by the LLM. 

\paragraph{Problem Setup}
The input consists of two parts: a document and multiple queries. The document is processed only by the \emph{Attention Graph Construction} stage, while the queries are handled by the \emph{Dynamic Graph Search} stage. 
Initially, a graph $\gG = (\gV, \gE)$ is constructed from a document, where $\gV$ denotes the collection of nodes and $\gE$ denotes the collection of edges. 
Each node $v_{i}^{l} \in \gV$ contains the text of an Information Point (IP) and belongs to level $l$. 
Each edge $e_{i,j} \in \gE$ indicates the relatedness between node $v_i$ and node $v_j$, derived from LLM attention weights. 
Each $v_i$ contains tokens $\{x_n^i\}_{n=1}^{N_i}$, where $N_i$ denotes the number of tokens in node $v_i$. 
The \emph{Dynamic Graph Search} stage then uses the graph $\gG$ to generate a response for each query $q$.

\subsection{Attention Graph Construction}
\label{sec:summary_graph_construction}

A document is initially split into chunks of 300 tokens, which serve as the first-level nodes $\gV_1$. 
For each level, we iteratively summarize nodes from $\gV_l$ by batching them and feeding them into the LLM to obtain the higher-level nodes $\gV_{l+1}$, as illustrated in Figure~\ref{fig:sum}. 
Specifically, we select nodes for each batch by sequentially adding them until the total text content length exceeds a threshold $s$. 
These nodes are then input into the LLM. We prompt the LLM to generate IPs in the form of bullet points. 
The LLM prompt used is shown 
in Appendix~\ref{app:prompt_sum}, and an example of the generated IPs is shown in Appendix~\ref{app:ip_ex}.
We found that this bullet-point prompt ensures that each point contains only a single or a few events and remaining easy for the LLM to process.

The attention from a higher-level node $v_{i}^{l+1}$ to a lower-level node $v_{j}^{l}$ is averaged and then normalized across attention weights between nodes, yielding the edge weight $e_{i,j}$ from node $v_i$ to node $v_j$.
The value of $e_{i,j}$ quantifies the extent to which $v_{i}^{l+1}$ relies on or extracts information from $v_{j}^{l}$.
We focus only on attention weights between high-level and low-level nodes, omitting those within the same node. This approach emphasizes long-distance semantic relationships.
The computational process is elaborated below:

\paragraph{Extracting attention weights} During summarization, LLM attention weights are captured and averaged across attention heads and layers. 
In practice, these weights are iteratively accumulated during each layer's inference, thereby limiting memory usage.
This results in token-level attention \( e_{x^i,x^j} \) between tokens \( x^i \in v_i^{l+1} \) and \( x^j \in v_j^{l} \).

\paragraph{Aggregating token-level attention to node-level}
For each node $v_j^{l}$, the attention from all its tokens to token
$\{e_{x^i,x_n^j}\}_{n=1}^{N_j}$
is averaged
to obtain the token-to-node attention $e_{x^i,v_j^{l}}$. 
Similarly, for node $v_i^{l+1}$, token-to-node attention $\{e_{x_n^i,v_j^{l}}\}_{n=1}^{N_i}$ is averaged to obtain the final node-level edge weight 
$e_{v_i^{l+1},v_j^{l}}$, abbreviated as
$e_{i,j}$.

\paragraph{Normalization}
Finally, the edges are normalized as
$e_{i,j} = \frac{e_{i,j}}{\sum_{j} e_{i,j}}$,
ensuring that $\sum_{j} e_{i,j} = 1$.
If no direct attention exists between two nodes, the corresponding edge weight is set to zero.

\subsection{Dynamic Graph Search}
\label{sec:dynamic_graph_search}

The dynamic graph search process comprises two components: \emph{Dynamic Progress Control}, which provides dynamic LLM-guided control, and \emph{Graph Search}, which executes the search process.

\subsubsection{Dynamic Progress Control}
\label{sec:dynamic_control}

The process begins by initializing a visited set, denoted as $\gS \subset \gV$, with the top-level nodes of $\gV$, which is fed into an LLM. 
The LLM is prompted to determine whether the current set of visited nodes $\gS$ is sufficient to answer a given query.
The prompt asks the LLM, ``\emph{Can this question be answered by the following information?}'', followed by the query and visited nodes $\gS$.
(The complete prompt is shown
in Appendix~\ref{app:prompt_sea}.)
If the response is ``\texttt{No}'', the search continues, and the next node is retrieved.
The newly retrieved node is added to the visited set $\gS$, and this process repeats until the LLM responds with ``\texttt{Yes}.''
Throughout the search, all previous inputs, including the prompt, query, and visited nodes $\gS$, are cached using KV caching, as illustrated in Figure~\ref{fig:sea}.
This ensures that no additional computational resources are required.
Once the LLM responds with ``\texttt{Yes}'', the second turn of the prompt will ask the LLM to answer the query and the final answer is obtained.
We introduce two hyperparameters, stop patience $t_n$ and confidence threshold $t_p$, to balance effectiveness and efficiency in the search process.  
The stop patience parameter $t_n$ ends the search once the LLM has answered ``\texttt{Yes}'' $t_n$ times, similar to the concept of early stopping patience. We set $t_n = 1$ in the experiments and analyze how varying $t_n$ affects the trade-off in Section~\ref{sec:exp_dyn}.  
The confidence threshold $t_p$ influences the termination based on model certainty: let $p_{\texttt{Yes}}$ and $p_{\texttt{No}}$ be the normalized probabilities of ``\texttt{Yes}'' and ``\texttt{No}'', respectively. When the normalized probability $p_{\texttt{Yes}} > t_p$, the LLM halts the search.
Note that these two hyperparameters control the overall effectiveness–efficiency trade-off, whereas Dynamic Progress Control manages the query-level trade-off, which is a capability that previous methods lack.

This dynamic approach, termed \textbf{Dynamic Progress Control}, better allocates limited resources among different queries to achieve a better balance between effectiveness and efficiency (see Section~\ref{sec:exp_num_node} for analysis).
In contrast, previous methods typically relied on a fixed amount of retrieved content, which led to either computational waste or insufficient information retrieval.

\subsubsection{Graph Search}
\label{sec:graph_sea}

After identifying a set of IP nodes required to answer a query, the set of visited nodes $\gS \subset \gV$ is provided to the LLM using the prompt shown in Table~\ref{table:pp_sea}.
The attention weight between a visited node $v_i$ and the query $q$, denoted as $r_{i}$, is generally extracted following the method described in Section~\ref{sec:summary_graph_construction}.
(See Appendix~\ref{app:query_attn_comp} for more details.)
This weight indicates the degree of attention that node $v_i$ gives to the query $q$.

Specifically, for a node $v_i$, we define its predecessors as $\gP_i$. The intersection of $\gP_i$ and the set of visited nodes $\gS$ is then $\gQ_i = \gP_i \cap \gS$.
For each node $v_j \in \gQ_i$, the retrieval score $z_i$ for a node $v_i \notin \gS$ is computed as:
\begin{equation} 
\label{eq:app_retrieval_score}
z_i = \sum_{j \in \gQ_i} r_j e_{j,i}
\end{equation}
where $e_{j,i}$ represents the edge weight from node $v_j$ to node $v_i$.
This operation is performed for all successors of nodes in $\gS$.

This process can also be carried out via matrix multiplication. 
We define the adjacency matrix $\mE \in \mathbb{R}^{|\gV| \times |\gV|}$, where $\mE_{i,j} = e_{i,j}$ and $|\gV|$ denotes the number of nodes in $\gV$. 
Next, we construct the score vector $\vr \in \mathbb{R}^{|\gV|}$ such that $\vr_{i} = r_i$ if $v_i \in \gS$, and $\vr_{i} = 0$ otherwise.
The final score vector $\vz \in \mathbb{R}^{|\gV|}$ is computed as $\vz = \mE^T \vr$, where each entry of $\vz$ represents the corresponding node's score.
Finally, the embedding similarity is added to $\vz$ as the final score, and the node with the highest final score is retrieved. 
Notably, \OurName\ without embedding similarity only shows a slight performance degradation, whereas using only embedding similarity results in a larger performance decline (see the ablation study in Section~\ref{sec:abl_exp}).

The key idea is that if a node (i.e., an Information Point) containing an event is strongly correlated with the query, then the details about that event are likely to be more useful in answering the query.
We use $r$ to represent the relevance of a high-level node to the query and $e$ to represent the relevance of a lower-level node to its associated high-level node.
A high $e$ indicates that the lower-level node contains more detailed information about the same event, as illustrated in Figures~\ref{fig:sum} and \ref{fig:ex_ho}.
If a retrieved node provides details that are not relevant to the query, its $r$ score will be low, preventing the search from continuing along that node.
If multiple nodes are highly relevant to both the query and the same successor, that successor node will accumulate scores from these multiple predecessors, resulting in a higher overall score.


\section{Experiments}


\begin{table*}[t]
\centering
\resizebox{0.85\linewidth}{!}{
\begin{tabular}{@{}lcccccccc@{}}
\toprule
\multirow{2}{*}{\textbf{Method}} & \multicolumn{4}{c}{\textbf{NarrativeQA}} & \multicolumn{4}{c}{\textbf{Qasper}} \\
\cmidrule(lr){2-5}\cmidrule(lr){6-9}
& \textbf{F1} & \textbf{ROUGE-L} & \textbf{TFLOPs} & \textbf{Ratio} & \textbf{F1} & \textbf{ROUGE-L} & \textbf{TFLOPs} & \textbf{Ratio} \\
\midrule
\textbf{BM25} Top-5 & 52.7 & 51.8 & 26.7 & 0.86x & 41.0 & 39.6 & 26.3 & 0.39x \\
\textbf{SBERT} Top-5 & 36.5 & 35.8 & 26.8 & 0.86x & 44.4 & 42.4 & 26.0 & 0.39x \\
\textbf{Dragon} Top-5 & 53.8 & 52.9 & 26.9 & 0.87x & 43.0 & 41.4 & 24.5 & 0.36x \\
\textbf{MeMWalker} & 11.2 & 9.8 & 353.8 & 11.41x & 39.0 & 36.8 & 123.9 & 1.85x \\
\textbf{RAPTOR-TT} & 40.6 & 39.8 & 20.3 & 0.65x & 42.1 & 40.1 & 17.7 & 0.26x \\
\textbf{RAPTOR-CT} Top-5 & 48.6 & 47.8 & 17.9 & 0.58x & 44.6 & 42.7 & 16.6 & 0.25x \\
\textbf{LongLLMLingua} & 50.5 & 49.5 & 1789.4 & 57.72x & 43.2 & 43.0 & 159.7 & 2.39x \\
\textbf{BM25} Top-$X$ & 53.7 & 52.9 & 37.5 & 1.21x & 47.0 & 45.1 & 69.3 & 1.04x \\
\textbf{SBERT} Top-$X$ & 39.5 & 38.8 & 37.5 & 1.21x & 46.6 & 44.5 & 68.9 & 1.03x \\
\textbf{Dragon} Top-$X$ & 55.1 & 54.2 & 37.5 & 1.21x & 46.9 & 44.8 & 67.0 & 1.00x \\
\textbf{RAPTOR-CT} Top-$X$ & 52.0 & 51.2 & 35.1 & 1.13x & 46.9 & 44.7 & 67.3 & 1.01x \\
\textbf{Llama-3.1-8B} & 53.7 & 52.6 & 3361.9 & 108.45x & 49.4 & 47.6 & 92.5 & 1.38x \\
\midrule
\textbf{\OurName} & \textbf{61.1} & \textbf{60.2} & 31.0 & 1.00x & \textbf{49.7} & \textbf{47.9} & 66.9 & 1.00x \\
\midrule
\multirow{2}{*}{\textbf{Method}} & \multicolumn{4}{c}{\textbf{HotpotQA}} & \multicolumn{4}{c}{\textbf{MuSiQue}} \\
\cmidrule(lr){2-5}\cmidrule(lr){6-9}
& \textbf{F1} & \textbf{ROUGE-L} & \textbf{TFLOPs} & \textbf{Ratio} & \textbf{F1} & \textbf{ROUGE-L} & \textbf{TFLOPs} & \textbf{Ratio} \\
\midrule
\textbf{BM25} Top-5 & 40.8 & 40.9 & 22.9 & 1.43x & 28.7 & 28.7 & 26.3 & 0.85x \\
\textbf{SBERT} Top-5 & 40.9 & 40.8 & 22.6 & 1.41x & 30.7 & 30.8 & 26.1 & 0.84x \\
\textbf{Dragon} Top-5 & 39.7 & 39.6 & 23.3 & 1.46x & 28.5 & 28.4 & 28.1 & 0.91x \\
\textbf{MeMWalker} & 39.7 & 38.9 & 93.4 & 5.84x & 24.0 & 23.5 & 175.7 & 5.69x \\
\textbf{RAPTOR-TT} & 38.6 & 38.5 & 8.4 & 0.53x & 29.3 & 29.3 & 12.6 & 0.41x \\
\textbf{RAPTOR-CT} Top-5 & 40.9 & 40.4 & 15.3 & 0.96x & 31.5 & 31.5 & 16.1 & 0.52x \\
\textbf{LongLLMLingua} & 43.4 & \textbf{43.5} & 43.6 & 2.73x & 34.5 & 34.4 & 78.9 & 2.55x \\
\textbf{BM25} Top-$X$ & 40.7 & 40.8 & 20.0 & 1.25x & 31.8 & 31.7 & 35.6 & 1.15x \\
\textbf{SBERT} Top-$X$ & 40.8 & 40.7 & 19.6 & 1.23x & 32.5 & 32.5 & 35.6 & 1.15x \\
\textbf{Dragon} Top-$X$ & 39.2 & 39.1 & 20.6 & 1.29x & 30.2 & 30.1 & 38.0 & 1.23x \\
\textbf{RAPTOR-CT} Top-$X$ & 40.7 & 40.7 & 17.9 & 1.12x & 35.4 & 35.2 & 32.2 & 1.04x \\
\textbf{Llama-3.1-8B} & 41.3 & 41.2 & 23.7 & 1.48x & 35.8 & 35.7 & 40.6 & 1.31x \\
\midrule
\textbf{\OurName} & \textbf{43.5} & \textbf{43.5} & 16.0 & 1.00x & \textbf{36.9} & \textbf{36.8} & 30.9 & 1.00x \\
\bottomrule
\end{tabular} }
\caption{\label{tab:main}
    F1 (\%), ROUGE-L (\%), and TFLOPs of baselines and \OurName\ on NarrativeQA, Qasper, HotpotQA, and MuSiQue. TFLOPs are calculated during query-dependent inference. ``Ratio'' represents the ratio of the baselines' TFLOPs to \OurName's TFLOPs. 
    In addition to Top-5, we also include a Top-$X$ setting to match the TFLOPs of \OurName\ for a fair comparison.
    For BM25, SBERT, and Dragon we used Top-7 (NarrativeQA), Top-14 (Qasper), Top-4 (HotpotQA), and Top-7 (MuSiQue); for RAPTOR-CT, we used Top-20, Top-42, Top-12, and Top-22, respectively.
}
\end{table*}

\begin{table*}[t]
\centering
\resizebox{0.8\linewidth}{!}{
\begin{tabular}{@{}cccccc@{}}
\toprule
\textbf{Graph Construction} & \textbf{Graph Search} & \textbf{NarrativeQA} & \textbf{Qasper} & \textbf{HotpotQA} & \textbf{MuSiQue} \\
\midrule
Llama-3.1-8B & Llama-3.1-8B & 61.1 & 49.7 & 43.5 & 36.9 \\
\midrule
Llama-3.2-3B & Llama-3.2-3B & 56.7 & 47.6 & 40.0 & 29.8 \\
Llama-3.1-8B & Llama-3.2-3B & 60.6 & 49.6 & 40.8 & 31.3 \\
\midrule
Llama-3.2-1B & Llama-3.2-1B & 28.8 & 33.5 & 27.5 & 15.3 \\
Llama-3.1-8B & Llama-3.2-1B & 41.8 & 37.2 & 28.5 & 15.6 \\
\bottomrule
\end{tabular} }
\caption{\label{tab:small_model}
    Comparison of different model combinations for graph construction and graph search, evaluated on NarrativeQA, Qasper, HotpotQA, and MuSiQue using F1~(\%).
}
\end{table*}


\paragraph{Dataset} 
We use two single-document QA and two multi-document QA datasets from LongBench \citep{bai-etal-2024-longbench}: 
\textbf{NarrativeQA} \citep{kocisky-etal-2018-narrativeqa} is a single-doc QA dataset containing 1,567 stories, including full texts of books and movie transcripts. 
\textbf{Qasper} \citep{dasigi-etal-2021-dataset} is a single-doc QA dataset with 1,585 papers, designed to seek information present in the papers. 
\textbf{HotpotQA} \citep{yang2018hotpotqa} is a multi-doc QA dataset that contains 112,779 examples, focusing on multi-hop QA. 
\textbf{MuSiQue} \citep{trivedi2021musique} is a multi-doc QA dataset with 24,814 examples featuring 2-4 hop questions and six reasoning types. 
See Appendix~\ref{app:datasets} for more details.

\paragraph{Metrics} We use F1 and ROUGE-L \citep{lin-2004-rouge} 
as evaluation metrics. 
The final scores are computed using the evaluation source code from LongBench \citep{bai-etal-2024-longbench} and Hugging Face Evaluate\footnote{\url{https://github.com/huggingface/evaluate}}. We also measure the average query-dependent TFLOPs (Tera Floating Point Operations) consumed during search and inference for each query.

\paragraph{Baseline} 
We employ the following baselines:
\textbf{BM25} \citep{robertson1995okapi,bm252009} is a bag-of-words based retrieval method
that ranks documents based on the query terms appearing in them.
\textbf{SBERT} \citep{reimers-gurevych-2019-sentence} is a dense retrieval method that employs dense embeddings obtained through the encoder model.
\textbf{Dragon} \citep{lin-etal-2023-train} 
uses contrastive learning and data augmentation to train a model, achieving state-of-the-art retrieval performance among eight RAG baselines.
\textbf{LongLLMLingua} \citep{jiang-etal-2024-longllmlingua} is a prompt compression method that introduces question-aware compression based on LLMLingua \citep{jiang-etal-2023-llmlingua}.
\textbf{MeMWalker} \citep{chen2023walkingmemorymazecontext} summarizes context into a tree and navigates it to search for relevant information guided by the LLM within a limited input window.
\textbf{RAPTOR} \citep{sarthi2024raptor} constructs a tree by recursively embedding, clustering, and summarizing chunks of text. 
It has two variants: ``tree traversal'' (\textbf{RAPTOR-TT}) retrieves nodes along a single path from the top of the tree to the bottom, and ``collapsed tree'' (\textbf{RAPTOR-CT}) flattens all nodes for standard RAG-based retrieval.
\textbf{Llama-3.1-8B} \citep{dubey2024llama3herdmodels} expands the context window, enabling documents to be fed into the model, except for a few from NarrativeQA.

We use Llama-3.1-8B \citep{dubey2024llama3herdmodels} as the LLM for \OurName\ and all baselines by running their source code.
For MeMWalker, since the source code was not released, we implemented it based on the description in the paper using Llama-3.1.
For LongLLMLingua, which employs a smaller model to compress prompts for GPT-3.5, we used the Phi-2-2.7B model provided by LongLLMLingua for compression.
For all methods, we adopt a zero-shot approach without Chain-of-Thought (CoT).
We use SBERT \citep{reimers-gurevych-2019-sentence} as the retrieval model in \OurName. 
We set 
the length threshold $s$ to 8K.
Across all datasets and steps of \OurName, including graph construction and search, the input window is capped by $s$ and can be processed using an NVIDIA A100 80G GPU.
A summary graph example is illustrated in Appendix~\ref{app:ip_ex}.


\begin{table*}[t]
\centering
\resizebox{0.9\linewidth}{!}{
\begin{tabular}{@{}lcccccccc@{}}
\toprule
\multirow{2}{*}{\textbf{Method}} & \multicolumn{2}{c}{\textbf{NarrativeQA}} & \multicolumn{2}{c}{\textbf{Qasper}} & \multicolumn{2}{c}{\textbf{HotpotQA}} & \multicolumn{2}{c}{\textbf{MuSiQue}} \\
\cmidrule(lr){2-3}\cmidrule(lr){4-5}\cmidrule(lr){6-7}\cmidrule(lr){8-9}
& \textbf{F1} & \textbf{ROUGE-L} & \textbf{F1} & \textbf{ROUGE-L} & \textbf{F1} & \textbf{ROUGE-L} & \textbf{F1} & \textbf{ROUGE-L} \\
\midrule
\textbf{\OurName} & 61.1 & 60.2 & 49.7 & 47.9 & 43.5 & 43.4 & 36.9 & 36.9 \\
\midrule
w/o Attention-Guided Retrieval & 53.0 & 52.4 & 46.9 & 45.5 & 41.9 & 41.7 & 33.1 & 32.9 \\
w/o Dynamic Progress Control & 53.5 & 52.9 & 37.7 & 36.4 & 39.4 & 39.3 & 27.0 & 27.1 \\
w/o IP-based Graph & 51.4 & 50.8 & 47.3 & 45.5 & 40.9 & 40.7 & 32.3 & 32.3 \\
w/o Embedding Similarity & 59.5 & 58.7 & 48.0 & 46.2 & 42.9 & 42.8 & 35.9 & 35.8 \\
\bottomrule
\end{tabular} }
\caption{\label{tab:abl}
Ablation study of the four components of \OurName\ with F1 (\%) and ROUGE-L (\%) on NarrativeQA, Qasper, HotpotQA, and MuSiQue.
}
\end{table*}


\subsection{Main Results}
\label{sec:exp_main}

The main results are shown in Table~\ref{tab:main}. 
In addition to Top-5 retrieval, we also include a Top-$X$ setting to match the TFLOPs of \OurName\ for a fair comparison.
For MeMWalker, RAPTOR, and \OurName, summarization TFLOPs are not included as summarization is query-independent. 
It is worth noting that our \emph{Dynamic Progress Control} can determine the appropriate number of nodes to retrieve in a single pass, whereas these methods require extensive hyperparameter searches to find the optimal number.
For Llama-3.1 on NarrativeQA, we used 8 A100 80G GPUs, with CPU offloading, to handle the long documents. However, we could only process an input window of 100K tokens under these resource constraints, resulting in 22.3\% of documents being truncated.

The performance of BM25, SBERT, and Dragon was similar, with Dragon showing an advantage on NarrativeQA.
When comparing the Top-5 and Top-$X$ results, retrieving more fragments generally leads to better performance.
LongLLMLingua achieves better results than Llama-3.1 on HotpotQA, possibly because it reorders documents to place the most relevant content upfront, mitigating the lost-in-the-middle effect \citep{liu-etal-2024-lost}. 
However, for other datasets, the deletion of sentences and tokens in LongLLMLingua negatively impacts its performance. 
The small size difference between Phi-2 and Llama-3.1-8B may not reflect the intended use cases of LongLLMLingua, making efficiency comparisons challenging.

MeMWalker requires the LLM to generate correct responses and formats at every node, which increases computational complexity and raises the risk of navigation failures when the tree becomes large.
This contributed to its poor performance on NarrativeQA.
RAPTOR-TT shows relatively poor performance as it is constrained by a fixed retrieval path from the top to the bottom.
RAPTOR-CT Top-$X$ achieves high performance but still underperforms compared to \OurName\ at comparable TFLOPs, suggesting that its tree-based summarization is less efficient. 
Llama-3.1-8B excelled on most datasets, demonstrating its strong capability. 
However, for particularly long inputs in NarrativeQA, Llama-3.1 incurred 108.45x TFLOPs, which is significantly more computationally expensive than using \OurName.

\OurName\ achieves a good balance between effectiveness and efficiency, and with fewer TFLOPs, 
outperforming all baselines across all four datasets in terms of F1 and ROUGE-L. 
Even for Top-$X$ setting, with fewer TFLOPs, \OurName\ results are still better than those of RAPTOR-CT.
Compared to LongLLMLingua, our method achieved better performance with lower TFLOPs, although the differing application scenarios limit direct comparison.
\OurName\ slightly outperforms Llama-3.1 on Qasper, HotpotQA, and MuSiQue with fewer TFLOPs, and significantly surpasses Llama-3.1 on NarrativeQA while incurring much lower computational cost on a single GPU, demonstrating its capability to handle extremely long documents.
Since \OurName\ begins with query-related information, the most relevant content is presented first, mitigating the "lost-in-the-middle" effect. With IPs organized by events, \OurName\ can better track these events, especially as long-distance relationships tend to weaken.


\newcommand{\widthnumnode}{0.244}
\begin{figure*}[t]
    \centering
    \includegraphics[width=\widthnumnode\linewidth]{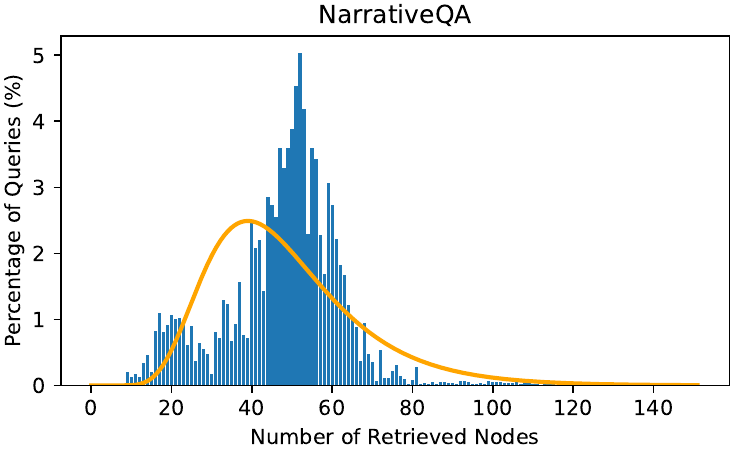}
    \includegraphics[width=\widthnumnode\linewidth]{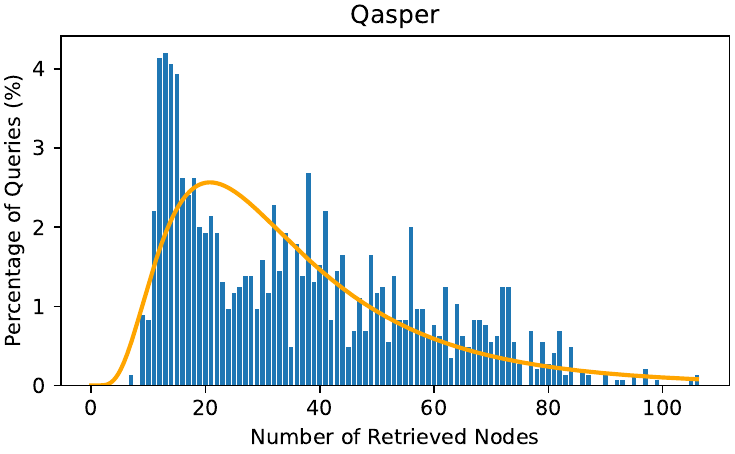}
    \includegraphics[width=\widthnumnode\linewidth]{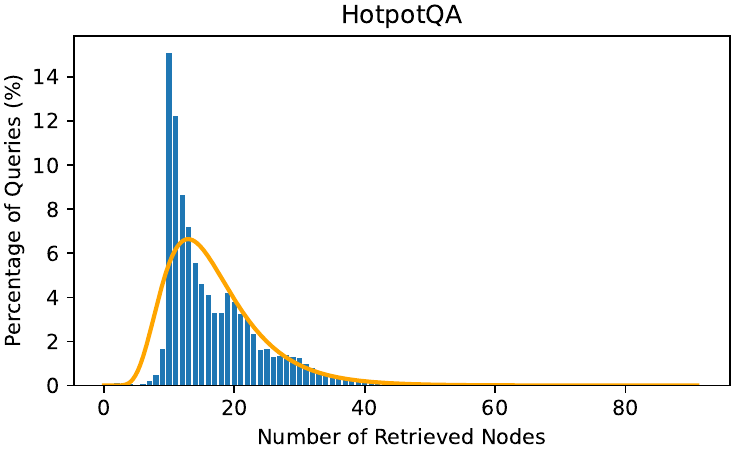}
    \includegraphics[width=\widthnumnode\linewidth]{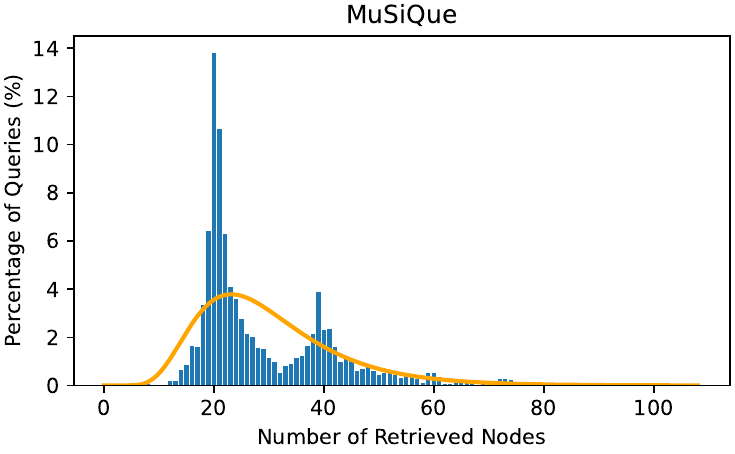}
    \caption{
        Frequency distribution of the number of nodes retrieved per query for NarrativeQA, Qasper, HotpotQA, and MuSiQue. The x-axis represents the number of nodes retrieved per query, while the y-axis indicates the percentage of queries retrieving that number of nodes. 
        A log-normal distribution is fitted, as shown by yellow line.
    }
    \label{fig:num_node}
\end{figure*}


\begin{figure*}[ht]
	\centering 
	\centerline{\includegraphics[width=0.99\textwidth]{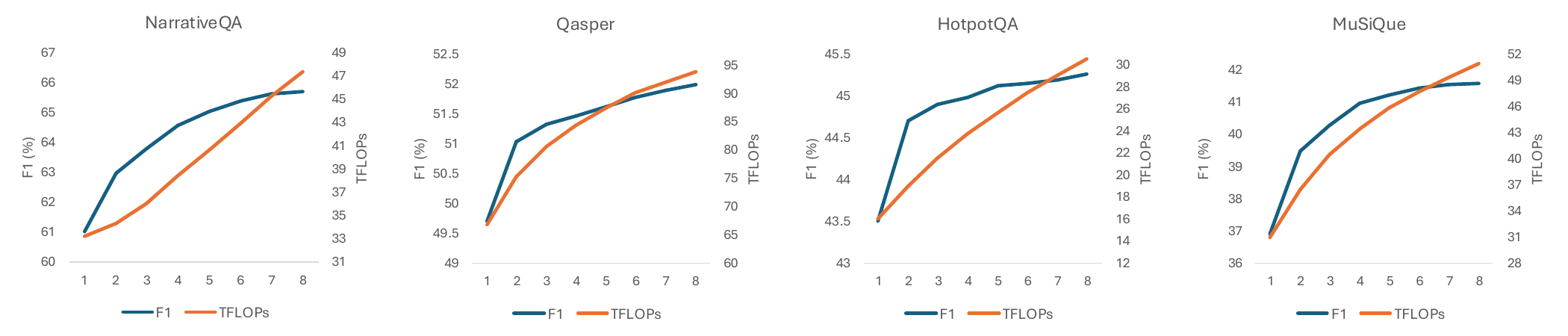}}
	\caption{
            The F1 (\%) and TFLOPs of \OurName\ on NarrativeQA, Qasper, HotpotQA, and MuSiQue with different stop patience $t_n$.
            The x-axis represents stop patience. The left y-axis shows the F1 (\%) (blue line), and the right y-axis shows TFLOPs (orange line).
            As $t_n$ increases from 1 to 8, both F1 and TFLOPs increase, but the increase in F1 slows when $t_n > 5$, while TFLOPs continue to rise.
        }
	\label{fig:stop} 
\end{figure*}

\paragraph{Inference with Smaller Models}
We further investigate whether the effectiveness-efficiency trade-off can be improved when the model used for graph search is smaller than the one used for graph construction.
We experiment with combinations of of Llama-3.1-8B, Llama-3.2-1B, and Llama-3.2-3B. 
The results are shown in Table~\ref{tab:small_model}.
For both Llama-3.2-3B and Llama-3.2-1B, employing a graph constructed by Llama-3.1-8B significantly improves performance compared with using graphs built by the smaller models themselves.
Remarkably, Llama-3.2-3B achieves performance comparable to Llama-3.1-8B when it leverages the graph generated by Llama-3.1-8B.


\subsection{Ablation Study}
\label{sec:abl_exp}

In this section, we study how each component contributes to performance, as shown in Table~\ref{tab:abl}, and describe them below.

\paragraph{w/o Attention-Guided Retrieval} 
We remove the use of attention 
and rely solely on embedding similarity for node search,
similar to RAPTOR. 
Performance dropped across all datasets, with the most significant drop occurring in NarrativeQA, indicating that attention scores are particularly effective in retrieving hierarchical information from long texts.

\paragraph{w/o Dynamic Progress Control} 
We conduct retrieval using a fixed number of nodes, corresponding to the average used by \OurName\ across the four datasets. 
Instead of employing \emph{Dynamic Progress Control}, we retrieve this fixed number of nodes. 
For documents with many top-level nodes, we limit the initial number of visited nodes $\gS$ to the preset value divided by the number of levels, with top-level nodes selected using embedding similarity as in RAPTOR to ensure sufficient exploration.
We retain the same search mechanism so that unselected top-level nodes can still be retrieved via embedding similarity. Without \emph{Dynamic Progress Control}, performance degraded across all datasets, even though the average number of nodes retrieved remained unchanged. 
See Section~\ref{sec:exp_num_node} for a deeper analysis.

\paragraph{w/o IP-based Graph} 
We do not use the IP-based many-to-many graph. Instead, following RAPTOR, we adopt a many-to-one summarization that ultimately forms a tree. 
Other search approaches remain unchanged. 
All datasets experience a performance decline, with NarrativeQA being particularly affected. This suggests that for complex and lengthy inputs, IPs with many-to-many graph-based relationships are more effective at organizing information and tracking events.

\paragraph{w/o Embedding Similarity}
We remove embedding similarity from the final score $z$, relying entirely on attention weights to compute retrieval scores. 
Across all four datasets, performance shows a slight decrease, though this drop is much smaller than that observed when the attention-guided retrieval is removed, indicating that attention-guided search plays a more critical role.


\subsection{Dynamic Retrieval Analysis}
\label{sec:exp_num_node}

In this section, we analyze the number of nodes retrieved by \emph{Dynamic Progress Control}. 
Since each node primarily contains an IP with only a small amount of text, \OurName\ can retrieve more nodes using the same TFLOPs.
NarrativeQA consists of stories, where each IP can be described succinctly, whereas Qasper comprises scientific papers, with IPs that are more complex and require longer descriptions.
Figure~\ref{fig:num_node} presents the frequency distribution of nodes retrieved per query for NarrativeQA, Qasper, HotpotQA, and MuSiQue. 
Most queries retrieve a moderate number of nodes, while a few require significantly more, forming a long tail. 
\OurName\ dynamically tailors the amount of retrieved information to accommodate this long tail, which is a typical scenario that previous methods cannot handle, as they typically retrieve a fixed number of node chunks. 
HotpotQA exhibits an overall right-skewed distribution. 
Qasper, which comprises 1,451 instances compared to HotpotQA’s 7,405, shows a wider range of retrieved nodes (mostly within 80, versus mostly within 30 for HotpotQA), resulting in greater noise in Qasper. 
NarrativeQA displays a small peak at a low number of retrieved nodes, potentially corresponding to easier queries. This observation aligns with \citet{kocisky-etal-2018-narrativeqa}'s note that ``a small number of questions and answers are shallow paraphrases of sentences in the full document.'' 
MuSiQue explicitly categorizes questions by the number of hops, with the majority being 2-hop questions and some requiring 3 or 4 hops, with fewer hops generally indicating a lower information requirement and vice versa. 
Figure~\ref{fig:num_node} also shows two distinct peaks, an observation that aligns with MuSiQue’s difficulty distribution and further demonstrates that \OurName\ can dynamically adapt to queries of varying complexity.


\subsection{Effectiveness-Efficient Trade-off Study}
\label{sec:exp_dyn}

In this section, we examine the trade-off between effectiveness and efficiency. 
As shown in Figure~\ref{fig:stop}, increasing $t_n$ can further enhance performance beyond the results reported in Table~\ref{tab:main}, albeit at the cost of increased computational resources. 
When $t_n < 5$, the F1 score rises rapidly, and for $t_n > 5$, the improvement in F1 slows while the TFLOPs increase disproportionately. 
All four datasets exhibit a similar trend with respect to $t_n$, demonstrating that \emph{Dynamic Graph Search} provides a good trade-off between effectiveness and efficiency. 
\OurName\ can adjust a single value of $t_n$ across all datasets to achieve a similar effectiveness-efficiency trade-off, whereas previous methods require hyper-parameter searches on each dataset separately.


\section{Conclusion}

In this paper, we introduced \OurName, a novel retrieval method that incorporates two key innovations. 
The LLM-Guided Dynamic Progress Control adjusts the amount of information retrieved based on the query, achieving a better balance between effectiveness and efficiency. 
The Attention-Guided Retrieval constructs a many-to-many Hierarchical Weighted Directed Acyclic Graph using LLM attention weights to guide the search. 
Each node represents an Information Point that focuses on one or a few events, enabling the model to effectively track them. 
Empirical results demonstrate that \OurName\ achieves LLM-level performance while maintaining RAG-level computational complexity.


\section*{Limitations}

\OurName\ primarily computes graph edges by averaging attention across all tokens, treating them as equally important. 
However, tokens may not actually be equally significant. 
Semantically irrelevant tokens (e.g., function words like ``a,'' ``an,'' ``the,'' etc.) might introduce noise. 
Nonetheless, since we only retain the attention between nodes and omit the attention within nodes, this can partially mitigate the issue. 
Moreover, tokens in different texts may appear in similar proportions, so the added noise is less noticeable. 
Nevertheless, our experiments show that this averaging approach still yields good results. 
We leave a deeper exploration of this issue, the development of a more fine-grained attention averaging strategy, for future work.
Similarly, we simply add the attention-based score and the embedding similarity score together, resulting in an equal weighting of both components. 
While this averaging approach also performs well in our experiments,
future work could explore when and to what extent each score should be weighted differently to achieve a more refined strategy. 
In this paper, we focus on presenting the main framework of Dynamic Progress Control and Attention-Guided Retrieval, leaving further refinements for future work.


\section*{Acknowledgements}

This work was supported in part by the UK Engineering and Physical Sciences Research Council (EPSRC) through a Turing AI Fellowship (grant no. EP/V020579/1, EP/V020579/2), a New Horizons grant (grant no. EP/X019063/1), KCL’s Impact Acceleration Account (grant no. EP/X525571/1).

\bibliography{custom}

\clearpage
\appendix

\setcounter{table}{0}
\renewcommand{\thetable}{A\arabic{table}}

\setcounter{figure}{0}
\renewcommand{\thefigure}{A\arabic{figure}}


\section{Prompt}
\label{app:prompt}

\subsection{Attention Graph Construction Prompt}
\label{app:prompt_sum}

In \emph{Attention Graph Construction}, we employ an LLM to generate Information Points (IPs) in a bullet-point format, with each bullet point representing an IP. 
Specifically, the LLM prompt is shown in Table~\ref{table:pp_sum}, along with an example from NarrativeQA \citep{kocisky-etal-2018-narrativeqa}. 
Each IP typically consists of a single sentence that describes one or a few events. T
his approach enables \emph{Graph Search} to more effectively retrieve information by tracking events rather than continuous text. 
We found that instructing the LLM to produce bullet-point lists aligns well with the LLM's inherent knowledge.

\begin{table*}[ht]
    \centering
    \resizebox{0.95\textwidth}{!}{
        \begin{tabular}{p{15cm}}
        \toprule
        \multicolumn{1}{c}{\textbf{Attention Graph Construction}} \\
        \midrule
        \textbf{Prompt:} \newline
        \newline
        Summary the following information. Each segment is separated by a new line symbol. \newline
        \newline
        \textit{......} \newline
        \textit{* Mrs. Tabitha Twitchit expects ``fine company'' for tea and fetches the children before her friends arrive.}\newline
        \textit{* Tabitha dresses Moppet and Mittens in clean pinafores and tuckers, and Tom in "all sorts of elegant uncomfortable clothes" taken from a chest of drawers.}\newline 
        \textit{* Tom is fat and bursts several buttons, but his mother sews them back on again.}\newline 
        \textit{* Tabitha turns her kittens into the garden to keep them out of the way while she makes hot buttered toast for the party.}\newline
        \textit{......} \newline
        \newline
        Split your summary into different summary points according to the semantic information in these information points. It is not necessary to generate each summary point for each information point. Gather and organize information into summary points. In each summary point, try to avoid using pronouns like he/she/they and instead use full names. Generate in the format of: \newline
        \newline
        * {summary point} \newline
        * {summary point} \newline
        * {summary point} \newline
        \textit{......} \newline
        \newline
        Do not provide any explanation and start the summary directly.
        \\
        \midrule
        \textbf{Response:} \newline 
        \newline
        \textit{* Mrs. Tabitha Twitchit expects fine company for tea and dresses Mittens, Tom Kitten, and Moppet in clean clothes.}\newline
        \textit{* Mrs. Tabitha Twitchit sends Mittens, Tom Kitten, and Moppet to the garden to keep them out of the way.}\newline
        \textit{......} \\
        \bottomrule
        \end{tabular}
    }
    \caption{
        The prompt used for the LLM during \emph{Attention Graph Construction}, along with an example from NarrativeQA. 
        The \textit{italicized} text represents the example context, while the remaining text represents the prompt instructions. 
        The prompt instructs the LLM to generate Information Points (IPs) in a bullet-point format, with each bullet point representing an IP, which aligns well with the LLM's inherent knowledge.
    }
    \label{table:pp_sum}
\end{table*}

\subsection{Graph Search Prompt}
\label{app:prompt_sea}

\emph{Graph Search} uses a two-turn prompt, as shown in Table~\ref{table:pp_sea}, with an example from NarrativeQA \citep{kocisky-etal-2018-narrativeqa}. 
In the first turn, the LLM is prompted to determine whether the current set of visited nodes $\gS$ is sufficient to answer the query. 
The prompt asks the LLM, ``\emph{Can this question be answered by the following information?}'', followed by the query and the visited nodes $\gS$.
If the response is ``\texttt{No}'', the search continues and the next node is retrieved, with all previous context KV cached to avoid additional computation.
This process repeats until the LLM responds with ``\texttt{Yes}''. 
Once the LLM responds with ``\texttt{Yes}'', the second turn of the prompt asks the LLM to answer the query.

\begin{table*}[ht]
    \centering
    \resizebox{0.95\textwidth}{!}{
        \begin{tabular}{p{15cm}}
        \toprule
        \multicolumn{1}{c}{\textbf{Dynamic Graph Search}} \\
        \midrule
        \textbf{First-Turn Prompt:} \newline
        \newline
        Can this question be answered by the following information? Response ``Yes'' or ``No'' in one word. Do not provide any explanation.\newline
        \newline
        Question: \newline
        \textit{Where does the mother send her kittens to keep them out of the way while getting ready for the party?} \newline
        \newline
        Information: \newline
        \textit{......} \newline
        \textit{* Mrs. Tabitha Twitchit sends Mittens, Tom Kitten, and Moppet to the garden to keep them out of the way.} \newline
        \textit{* Tabitha turns her kittens into the garden to keep them out of the way while she makes hot buttered toast for the party.} \newline
        \textit{......}
        \\
        \midrule
        \textbf{First-Turn Response:} \newline
        \newline
        Yes
        \\
        \midrule
        \textbf{Second-Turn Prompt:} \newline
        \newline
        Given the above information and question, answer the question as concisely as you can. 
        \\
        \midrule
        \textbf{Second-Turn Response:} \newline
        \newline
        \textit{The garden.} \\
        \bottomrule
        \end{tabular}
    }
    \caption{
        Prompt used for \emph{Graph Search}, with an example from NarrativeQA. 
        The \textit{italicized} text represents the example context, while the remaining text represents the prompt instructions. 
        In the first turn, the LLM is asked whether the current context is sufficient to answer the query. 
        If the response is ``\texttt{No},'' additional context is appended to the prompt, and the query is repeated. 
        Once the LLM responds with ``\texttt{Yes},'' the second turn prompts the LLM to provide the final answer directly.
    }
    \label{table:pp_sea}
\end{table*}


\section{KV Caching in Dynamic Progress Control}
\label{app:kv_cache}

This section explains how the KV cache described in Section~\ref{sec:dynamic_control} is utilized in our approach. 
In conventional scenarios, KV caching is performed at the token level, where the LLM caches the key-value states of previous tokens when generating the next token. In contrast, our method employs a node-level KV cache.

Given the last retrieved node $v_i$ in the visited set $\gS$, let $\{x_n^i\}_{n=1}^{N_i}$ denote the tokens of node $v_i$, where $N_i$ represents the number of tokens in $v_i$.
The query and visited nodes are represented as $[q, v_1, \dots, v_i, \dots, v_{|S|}]$, with their corresponding tokens organized as $[x_1^q, \dots, x_{N_q}^q, x_1^1, \dots, x_{N_1}^1, \dots, x_1^i, \dots, x_{N_i}^i]$.

When a new node $v_j$ is retrieved, the query, key, and value states for its tokens $\{x_n^j\}_{n=1}^{N_j} \in v_j$ are computed using the LLM's self-attention mechanism. Since $v_j$ also attends to the previously visited nodes, the stored KV cache containing the tokens $[x_1^q, \dots, x_{N_q}^q, x_1^1, \dots, x_{N_1}^1, \dots, x_1^i, \dots, x_{N_i}^i]$ is provided as input to the LLM. At this point, the query states from $v_j$ and the key-value states from $q$, $\gS$, and $v_j$ are available for self-attention.

After processing, $v_j$ is added to the visited set $\gS$, and the key-value states of its tokens $\{x_n^j\}_{n=1}^{N_j}$ are stored. These states are concatenated with the previous KV cache to form $[x_1^q, \dots, x_{N_q}^q, \dots, x_1^i, \dots, x_{N_i}^i, x_1^j, \dots, x_{N_j}^j]$, which is then used for the subsequent node retrieval. 

Once the retrieval process is complete, the key-value states of all visited nodes in $\gS$ are cached, and the LLM is prompted to answer the question using these cached states. 
Some LLMs may insert special tokens between the prompt and response, but these tokens are minimal, and the additional computation is negligible. 
The LLM follows standard decoding to generate tokens sequentially, leveraging the KV cache from previous tokens. Importantly, the query, key, and value states for all nodes, the query, and the answers are computed only once throughout the retrieval process, thereby avoiding additional computational overhead.


\section{Query-Node Attention Computation}
\label{app:query_attn_comp}

This section describes our method for computing the query-node relevance $r$, which is derived from the attention weights introduced in Section~\ref{sec:graph_sea}.

We treat the query as a node and employ the averaging method described in Section~\ref{sec:summary_graph_construction} to compute the relevance score $r_i$ between the query and each node $v_i$. 
Similarly, we extract only the attention between the node and the query, omitting other attention components such as intra-node attention. 
We do not apply the normalization procedure from Section~\ref{sec:summary_graph_construction}, which constrains the relevance scores to sum to 1, because a query may be related to multiple nodes, and different queries can be associated with varying numbers of nodes. 
As a result, each node may potentially exhibit a strong relation to the query.

Instead, we adopt a heuristic normalization approach. Since the query is positioned before the visited nodes in the prompt, nodes appearing later may allocate some of their attention to earlier nodes, often resulting in lower attention scores for later nodes. 
Although placing the query at the end would avoid this issue, it would prevent the query from being cached in the key-value memory, thereby creating a trade-off between performance and efficiency. 
To address this, we apply an empirical normalization to $r_i$ by multiplying it by the position index of $v_i$, with the query $q$ occupying the first position. For instance, in Figure~\ref{fig:sea}, the extracted attention $r_{15}$ is multiplied by 4 because it corresponds to the fourth position.
Note that this is merely a heuristic approach rather than an ultimate solution. 
In this paper, we focus on the primary concept of leveraging attention weights and leave a more in-depth exploration of this detail to future work.



\begin{figure*}[t]
	\centering 
        \centerline{\includegraphics[width=0.99\textwidth]{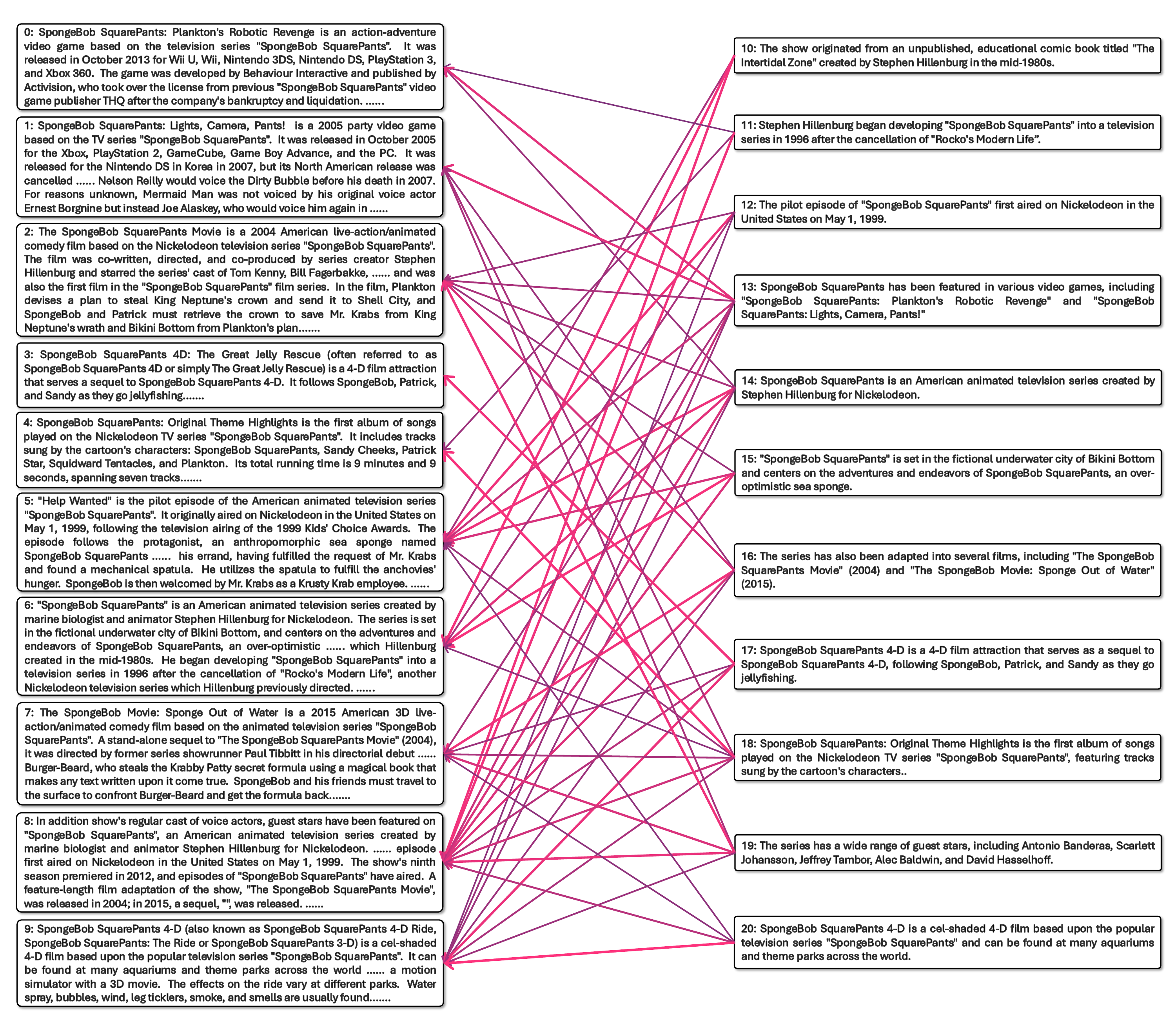}}
	\caption{
            An example of generated IPs with portions of the first- and second-level nodes from HotpotQA. The nodes on the right are second-level nodes generated from the first-level nodes on the left. The connections in the middle represent attention weights, with higher weights shown as redder and thicker lines. For brevity, lines with attention weights less than 0.05 are omitted, and some lengthy text is truncated.
        }
	\label{fig:ex_ho} 
\end{figure*}

\section{Summary Graph Example}
\label{app:ip_ex}

In this section, we present an example of generated Information Points (IPs) using parts of the first- and second-level nodes from HotpotQA \citep{yang2018hotpotqa} (see Figure~\ref{fig:ex_ho}). The nodes on the right represent second-level IPs, which are generated by aggregating information from the first-level chunk nodes on the left. The connections between these nodes denote attention weights, with higher weights visualized as redder and thicker lines.

It can be observed that the IPs are derived from multiple chunks. For instance, second-level nodes 10, 13, 14, 15, 16, 17, 18, and 19 are connected to several first-level nodes, while second-level nodes 11 and 20 primarily rely on a single first-level node with minimal connections elsewhere. From the perspective of the first-level nodes, nodes 2, 5, 6, 7, 8, and 9 are connected to multiple second-level nodes, whereas first-level node 3 is attended to only by second-level node 17.

The attention weights illustrate the dynamic connectivity between nodes. Some nodes connect to multiple others, while others connect to only one.
Overall, our method leverages IPs and attention to explicitly capture the relationships between first-level and second-level nodes, enabling the model to understand how each piece of information is interconnected within the text.


\section{Datasets}
\label{app:datasets}

\begin{table*}[ht]
\centering
\resizebox{0.6\linewidth}{!}{
\begin{tabular}{lccc}
\toprule
\textbf{Dataset} & \textbf{Average} & \textbf{Min} & \textbf{Max} \\
\midrule 
\textbf{NarrativeQA} \citep{kocisky-etal-2018-narrativeqa} & 79,457	& 5,077 & 467,867 \\
\textbf{Qasper} \citep{dasigi-etal-2021-dataset} & 4,866	& 918 & 29,408 \\
\textbf{HotpotQA} \citep{yang2018hotpotqa} & 1,318 & 70 & 3,575 \\
\textbf{MuSiQue} \citep{trivedi2021musique} & 2,267	& 909 & 4,432 \\
\bottomrule
\end{tabular}
}
\caption{Token length statistics for the NarrativeQA, Qasper, HotpotQA, and MuSiQue datasets, including the average, minimum, and maximum token lengths per document.}
\label{table:dataset_st}
\end{table*}

Table~\ref{table:dataset_st} shows the token length statistics for NarrativeQA \citep{kocisky-etal-2018-narrativeqa}, Qasper \citep{dasigi-etal-2021-dataset}, HotpotQA \citep{yang2018hotpotqa}, and MuSiQue \citep{trivedi2021musique}. 
In particular, NarrativeQA is much longer than the other datasets, followed by Qasper, while HotpotQA and MuSiQue contain relatively shorter texts. 
The confidence threshold $t_p$ is set to $0.5$, $0.98$, $0.5$, $0.55$ for NarrativeQA, Qasper, HotpotQA, and MuSiQue, respectively. 
We observe that the LLM is particularly confident on Qasper when deciding whether sufficient evidence has been gathered. 
As a result, it often terminates prematurely, yielding lower task performance and extremely low TFLOPs, which complicates fair comparison with other baselines.
However, by tuning $t_p$ we can raise both performance and computational cost, enabling comparisons under comparable compute budgets.
Qasper consists of scientific papers and requires precise, detailed answers, whereas \OurName\ often retrieves vague, summary-level information at first, which persuades the LLM that the query has been resolved and causes the search to terminate prematurely.
Note that the number of retrieved nodes still varies across documents.


\section{Efficiency Analysis}
\label{app:efficiency_analysis}

\begin{table*}[ht]
\centering
\resizebox{0.85\linewidth}{!}{
\begin{tabular}{@{}lcccc@{}}
\toprule
\multirow{2}{*}{\textbf{Method}} & \textbf{NarrativeQA}   & \textbf{Qasper}   & \textbf{HotpotQA}   & \textbf{MuSiQue} \\ 
& \textbf{TFLOPs}   & \textbf{TFLOPs}   & \textbf{TFLOPs}   & \textbf{TFLOPs} \\ 
\midrule
\textbf{Llama-3.1-8B} & 3361.9 & 92.5 & 23.6 & 40.6 \\
\midrule
\textbf{\OurName} Attention Graph Construction & 2042.8 & 136.6 & 40.2 & 66.7 \\ 
\textbf{\OurName} Graph Search & 31.0 & 66.9 & 16.0 & 30.9 \\ 
\midrule
\textbf{\OurName} Attention Graph Construction + Graph Search &  &  &  & \\ 
\quad 1 queries per document & 2073.8 & 203.5 & 56.2 & 97.6 \\ 
\quad 2 queries per document & 1052.4 & 135.2 & 36.1 & 64.3 \\ 
\quad 4 queries per document & 541.7 & 101.1 & 26.1 & 47.6 \\ 
\quad 8 queries per document & 286.4 & 84.0 & 21.0 & 39.2 \\ 
\bottomrule
\end{tabular} 
}
\caption{
    TFLOPs for graph construction and graph search. \OurName\ \emph{graph construction + graph search} shows the average TFLOPs per query when each document contains 1, 2, 4, or 8 queries.
}
\label{tab:eff}
\end{table*}

Table~\ref{tab:eff} presents the TFLOPs for graph construction per document (\OurName\ \emph{Attention Graph Construction}) and for graph search per query (\OurName\ \emph{Graph Search}). The combined TFLOPs of \OurName\ (\emph{Attention Graph Construction} + \emph{Graph Search}) represent the average TFLOPs per query for documents containing 1, 2, 4, or 8 queries.

As the number of queries per document increases, the TFLOPs for graph construction are amortized over more queries, reducing the average TFLOPs per query. In fact, when a document contains more than 8 queries, our method achieves lower average TFLOPs per query, even after accounting for summary generation.

For documents with only one query, our method's TFLOPs exceed those of Llama-3.1 on Qasper, HotpotQA, and MuSiQue. 
During graph construction, \OurName\ processes the entire document along with additional summary generation. 
However, on long documents such as those in NarrativeQA, \OurName\ uses fewer TFLOPs than Llama-3.1, since the Transformer \citep{attention_is_all_you_need} incurs quadratic complexity with very long inputs.
Moreover, while Llama-3.1 processes the entire input at once, our method processes the document in chunks. As a result, \OurName\ operates with an 8K input window, enabling it to run on a single GPU. In contrast, running Llama-3.1 on NarrativeQA with a 100K-token input window required 8 A100 80G GPUs (even with CPU offloading). This limitation hinders the practical application of Llama-3.1 in real-world scenarios, whereas our method can easily run on a single GPU.

\end{document}